\documentclass[sigconf]{acmart}
\usepackage{booktabs}
\usepackage[ruled,lined]{algorithm2e}
\usepackage{natbib}
\usepackage{mathtools}
\usepackage{graphicx}
\usepackage{subfigure}
\usepackage{algorithmic}

\usepackage{xcolor}

\copyrightyear{2022} 
\acmYear{2022} 
\setcopyright{rightsretained} 
\acmConference[GECCO '22]{Genetic and Evolutionary Computation Conference}{July 9--13, 2022}{Boston, MA, USA}
\acmBooktitle{Genetic and Evolutionary Computation Conference (GECCO '22), July 9--13, 2022, Boston, MA, USA}
\acmDOI{10.1145/3512290.3528718}
\acmISBN{978-1-4503-9237-2/22/07}

\newcommand{\fcnneme}{DSA-ME}
\newcommand{\lineareme}{LSA-ME}
\newcommand{\fixedfcnneme}{Offline DSA-ME}
\newcommand{\fcnnememt}{DSA-ME (ad)}

\newcommand{\fixedfcnnememt}{Offline DSA-ME (ad)}
\newcommand{\mapelites}{MAP-Elites}

\begin{document}

\title{Deep Surrogate Assisted MAP-Elites  \\ for Automated Hearthstone Deckbuilding}
\author{Yulun Zhang}
\affiliation{%
  \institution{Viterbi School of Engineering\\University of Southern California}
  \city{Los Angeles} 
  \state{CA} 
    \country{USA}
}
\email{yulunzha@usc.edu}

\author{Matthew C. Fontaine}
\affiliation{%
  \institution{Viterbi School of Engineering\\University of Southern California}
  \city{Los Angeles} 
  \state{CA} 
    \country{USA}
}
\email{mfontain@usc.edu}

\author{Amy K. Hoover}
\affiliation{%
  \institution{Ying Wu College of Computing\\New Jersey Institute of Technology}
  \city{Newark} 
  \state{NJ} 
    \country{USA}
}
\email{ahoover@njit.edu}

\author{Stefanos Nikolaidis}
\affiliation{%
  \institution{Viterbi School of Engineering\\University of Southern California}
  \city{Los Angeles} 
  \state{CA} 
    \country{USA}
}
\email{nikolaid@usc.edu}

\begin{abstract}

We study the problem of efficiently generating high-quality and diverse content in games. Previous work on automated deckbuilding in Hearthstone shows that the quality diversity algorithm \mbox{MAP-Elites} can generate a collection of high-performing decks with diverse strategic gameplay. However, MAP-Elites requires a large number of expensive evaluations to discover a diverse collection of decks. We propose assisting MAP-Elites with a deep surrogate model trained online to predict game outcomes with respect to candidate decks. MAP-Elites discovers a diverse dataset to improve the surrogate model accuracy, while the surrogate model helps guide MAP-Elites towards promising new content. In a Hearthstone deckbuilding case study, we show that our approach improves the sample efficiency of MAP-Elites and outperforms a model trained offline with random decks, as well as a linear surrogate model baseline, setting a new state-of-the-art for quality diversity approaches in automated Hearthstone deckbuilding. We include the source code for all the experiments at: \url{https://github.com/icaros-usc/EvoStone2}.  %

\end{abstract}

\maketitle 
\section{Introduction}

In this work, we study the problem of \textit{efficiently} generating high-quality and diverse content in games. We believe automated deckbuilding in collectable card games (i.e., Magic the Gathering~\cite{Magic_the_Gathering}, Legends of Runeterra~\cite{Legends_of_Runeterra}, or Hearthstone~\cite{hearthstone}) forms an ideal case study for this problem. In collectable card games, designers want to automatically generate high performing decks with diverse strategies for both player experience~\citep{Heijne2017PCGZelda} and playtesting~\citep{garcia:kbs18}. Example strategies include \textit{aggro} decks that focus on reducing the opponent hero's health to zero as quick as possible, and \textit{control} decks that only deal damage to the opponent once the board is under control~\citep{fontaine:gecco19}.

In the Hearthstone deckbuilding problem, generating high-quality and diverse decks is challenging. To evaluate a candidate deck requires at least 200 games even against restricted opponents~\citep{hoover2019ai, bhatt:fdg18}, because of the large number of initial game states and stochastic player actions. In other words, the outcome of a single Hearthstone game has high variance even with deterministic strategies.  

To generate high-quality decks with diverse strategic gameplay, we build upon previous work~\citep{fontaine:gecco19} that formulates the problem of automated deckbuilding as a quality diversity (QD) problem and generates collections of decks via the QD algorithm MAP-Elites. A QD formulation requires an objective function to maximize and measure functions to span. In the context of Hearthstone, we specify the objective as winning games and the measure functions as the average game length and average cards in hand. MAP-Elites will then generate a diverse collection of decks that vary according to these measures. For example, the algorithm will produce decks that end games quickly as well as decks that extend the game as long as possible. It will also attempt to find the best-performing decks for each measure output combination, e.g., for different game lengths.

\begin{figure*}[!h]
  \centering
  \includegraphics[scale=0.1]{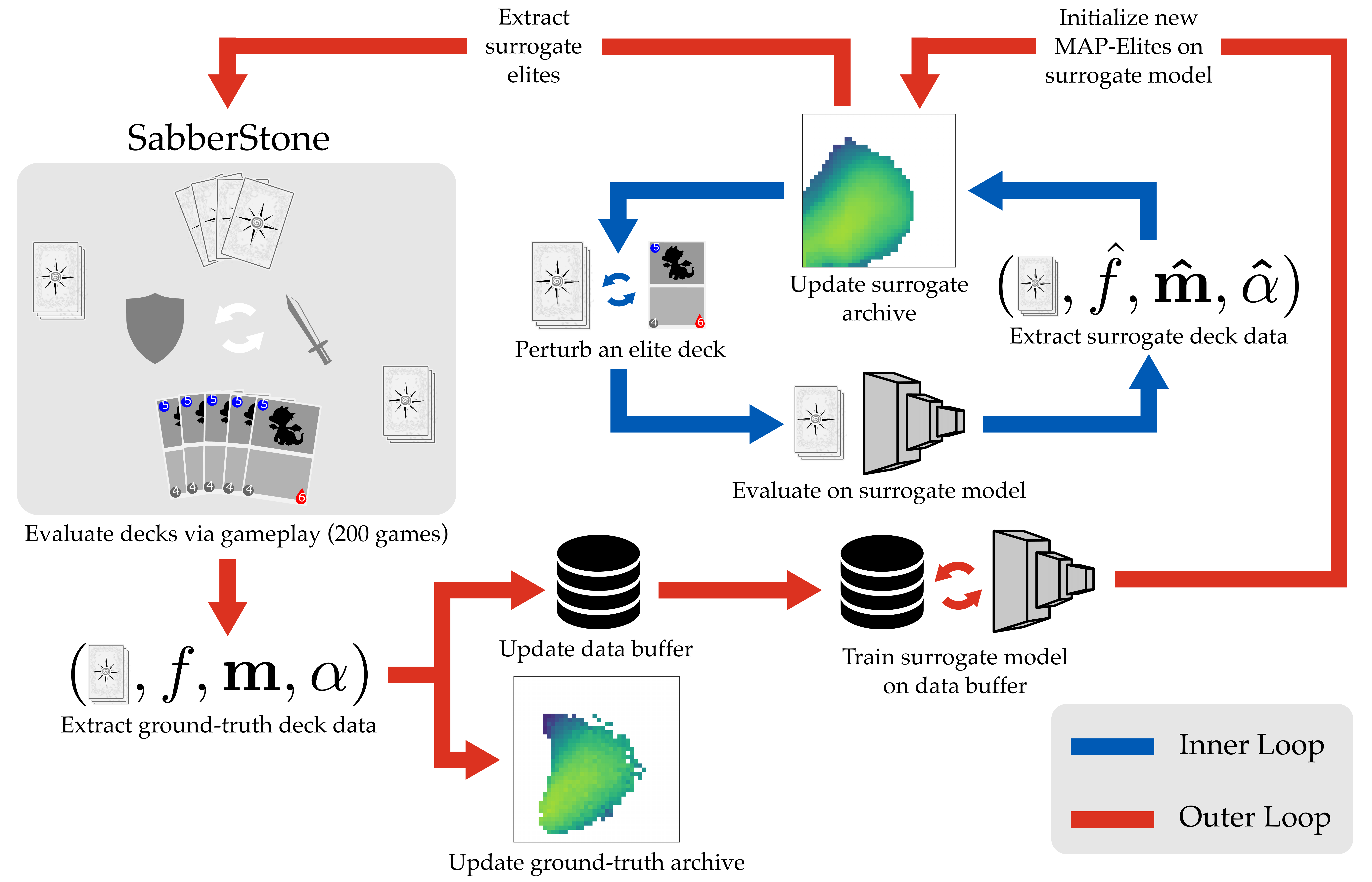}
  \caption{An overview of the Deep Surrogate Assisted MAP-Elites (DSA-ME) algorithm applied to Hearthstone Deckbuilding: Our approach consists of an inner loop (in blue) and an outer loop (in red). In the inner loop, we exploit a deep surrogate model of the SabberStone simulator by searching for a diverse collection of high performing decks with the MAP-Elites algorithm. Each iteration of the inner loop is a full run of the MAP-Elites algorithm on an input surrogate model. The elites of the MAP-Elites archive form candidate decks to be evaluated on the Hearthstone simulator \textit{SabberStone} in the outer loop. We perform an expensive evaluation of each deck by playing 200 games and extract ground-truth data about performance and playstyle. These candidate decks are both added to a data buffer as labeled data to train the surrogate model and to a ground-truth archive. After training the surrogate model on this new data, we initialize a new MAP-Elites to run in the inner loop on the updated surrogate model, completing one iteration of our algorithm. The result of DSA-ME is the ground-truth archive after several iterations of the outer loop.}
  \label{dsail}
\end{figure*}

However, searching for game content with MAP-Elites requires a large number of expensive evaluations, where each evaluation requires many runs in a game engine. We propose assisting the search with trained surrogate models that predict game outcomes for specific content. Previous work~\citep{Karavolos2021AMS} from procedural content generation showed how to train surrogate models offline from labeled gameplay data in a supervised fashion. Any search algorithm can then exploit the surrogate model directly on the game itself. However, a supervised surrogate model approach requires a large diverse labeled dataset and is vulnerable to the search discovering adversarial examples~\citep{trabucco2021conservative}, low-quality solutions that are predicted by the surrogate model to have high-quality.

As an alternative to offline surrogate models, we can acquire online labeled data by evaluation of game content with simulated agents. The Surrogate-Assisted Illumination (SAIL)~\cite{gaier2018dataefficient} algorithm trains on online surrogate model to improve the sample efficiency of MAP-Elites on generative design tasks. SAIL combines a surrogate model, modeled as a Gaussian process, that predicts the utility of solutions with an acquisition function that predicts where to sample. However, Gaussian processes can only be applied on low dimensional continuous data, whereas procedurally generated data is often high dimensional and discrete. For example, a Hearthstone cardset used to create decks may contain several hundred cards.

We propose training a deep neural network as a surrogate model that allows us to scale to high-dimensional search spaces. The deep neural network predicts both the objective and measures, thus we can run many iterations of MAP-Elites on a given surrogate model. Our method yields a two-fold benefit: \textit{by exploiting the surrogate model with \mbox{MAP-Elites}, we obtain diverse data to train a better surrogate model in future iterations and a better surrogate model helps guide MAP-Elites towards high-quality and diverse content.}

Overall, we introduce a novel model-based QD algorithm that maintains an online deep surrogate model predicting the objective and measures of generated game content. Our experiments in automated Hearthstone deckbuilding show that our approach improves the sample efficiency of MAP-Elites and outperforms a model trained offline, as well as an alternative linear model, setting a new state-of-the-art for quality diversity approaches to automated Hearthstone deckbuilding.

\section{Background}

\subsection{Hearthstone and Automated Deckbuilding} 

Hearthstone~\citep{hearthstone} is a virtual collectable card game published by Blizzard in 2014. Players need to form their own decks by choosing cards from a provided cardset. However, the game constrains deckbuilding by restricting player decks to contain exactly 30 cards and allows just two copies of each card with the exception of \textit{legendary} cards, which are restricted to one copy of each legendary per deck. During a match, two players compete directly against each other. At the start of the game, each player has 30 health points, each deck is shuffled into a random order, and the players draw their initial hand. Players take their turn by drawing a card, then playing cards from their hand with a specific mana cost. Cards either cast a spell that affects the game state or put a minion into play that can attack the opponent or other minions. Each player's objective is to decrease their opponent's health points to zero by their attacking with minions or casting damaging spells.

Due to the large cardinality of the Hearthstone deckspace, automated deckbuilding is considered as one of the two main AI challenges of Hearthstone~\citep{hoover2019ai}. At the time of this writing, Blizzard has published over 2500 collectible cards, which form a large search space of possible decks that players can design. We focus on the basic and classic card sets of the game, which are the first two sets of cards published by Blizzard in Hearthstone. Each of them contains 171 playable cards and each card can be added to a deck at most twice, which would create approximately $1.42\times10^{35}$ possible decks~\citep{fontaine:gecco19}. Although we only focus on these two initial sets, the size of the search space defined by them is large enough for us to demonstrate the effectiveness of our method.

Both evolutionary and non-evolutionary approaches have been applied to the automated deckbuilding problem. Evolutionary approaches include genetic algorithms~\citep{garcia:cig16, garcia:kbs18} and evolutionary strategies~\citep{bhatt:fdg18} that search the space of possible decks and find a single high-performing deck. Later work~\citep{fontaine:gecco19} applied the MAP-Elites~\citep{mouret2015illuminating} algorithm to the automated deckbuilding problem to generate a collection of high-performing decks with diverse strategic gameplay. Non-evolutionary approaches include a utility based system that automatically constructs decks based on predefined metrics~\citep{stiegler:skim16}, and a deckbuilding recommendation system~\citep{chen2018qdeckrec}.

In addition to deckbuilding, Hearthstone presents a unique challenge due to the stochasticity of initial state and actions, the large branching factor, the partial observability of the game state, and the large variety of possible opponents. As a result, many works train AI agents to play Hearthstone~\citep{ward:cig09,santos:cig17,stiegler:17,janusz:17,zhang:cig17,kachalsky:icmla17,dockhorn:ipmu18, swiechowski:cig18,choe:cog19,garcia:kbs20}. Other works predict the game result given a partial game log~\citep{bursztein:blog16,jakubik:fedcsis18} or predict the archetype of a deck using the first game round~\citep{eger:fdg20}.
Previous work~\citep{Fontaine_CMAME_2020} demonstrates that QD algorithms can generate a collection of Hearthstone agents with diverse strategies for a single deck.

\subsection{Quality Diversity Algorithms}

Quality diversity (QD) optimization~\citep{justin_qd_2016, chatzilygeroudis2021quality} is a class of stochastic optimization algorithms. We define the QD problem over a search space with an objective function to maximize -- known as a fitness function in evolutionary computation -- and measure functions\footnote{Prior works refer to the measure function outputs as behavior characteristics, behavior descriptors, features, or outcomes.} to span. Quality diversity algorithms have been proposed based on genetic algorithms~\cite{lehman2011abandoning, Cully_2015_robot_animal, lehman2011evolving}, evolution strategies~\citep{Fontaine_CMAME_2020, colas2020scaling, conti2017improving, nordmoen2018dynamic}, Bayesian optimization~\citep{kent2020bop}, and gradient ascent~\citep{fontaine2021differentiable}. The MAP-Elites~\citep{Cully_2015_robot_animal, mouret2015illuminating} QD algorithm pre-tessellates the space of measure function outputs. MAP-Elites then searches for a solution for each cell in the tessellation that maximizes the given objective. MAP-Elites returns an \textit{archive} of solutions; each cell of the archive contains at most one solution, i.e., an \emph{elite}, which is the highest performing solution in that cell.

As both the objective and measure functions are part of the problem definition, a user can apply QD to different domains by reducing their problem to a quality diversity problem. As a result, QD algorithms have been applied to procedural content generation~\cite{gravina2019procedural},  robotics~\citep{Cully_2015_robot_animal, mouret2015illuminating}, aerodynamic shape design~\citep{gaier2018dataefficient}, and scenario generation in human-robot interaction~\cite{fontaine2021quality,fontaine2021importance}.

\subsection{Model-based Quality Diversity Optimization}
Several works in quality diversity optimization have combined insights from model-based optimization (MBO)~\citep{bartz2016mobsurvey} or model-based reinforcement learning (MBRL)~\citep{Moerland2020ModelbasedRL}. These approaches broadly fit under the category of model-based quality diversity (MBQD). In MBO there are two main categories of methods: \textit{surrogate} methods and \textit{generative} methods.

\textit{Surrogate} model methods learn a predictive model of the optimization objective and then exploit that model through an acquisition function or direct optimization. In model-based quality diversity, the Surrogate-Assisted Illumination (SAIL) \citep{gaier2018dataefficient} algorithm and the Bayesian Optimization of Elites (\mbox{BOP-Elites}) algorithm~\citep{kent2020bop} both can be considered surrogate model methods as they learn a Gaussian process that is exploited via an acquisition function. While SAIL learns a surrogate model only of the objective, Surrogate-Assisted Phenotypic Niching (SPHEN)~\cite{hagg2020designing} uses Gaussian processes to predict both the objective and the measures. Previous work~\cite{cazenille2019exploring} has shown that MAP-Elites with surrogate models of micro-robots physics and chemical reaction dynamics resulted in performance comparable to a complete physical model.  Our proposed \mbox{DSA-ME} algorithm falls under this category of MBQD.

\textit{Generative} model methods learn a generative model of solutions as a learned representation of the search space. By creating a generative model of the search space, an optimization method can more efficiently search for complex solutions indirectly. From an evolutionary computation perspective, these methods can also be viewed as a type of indirect encoding of their search space~\citep{d2014hyperneat}. In model-based quality diversity, the Data-Driven Encoding MAP-Elites (DDE-Elites) algorithm~\citep{Gaier_dde_2020} and the Policy Manifold Search (PoMS) algorithm~\citep{rakicevic2021policy} both construct an online learned representation of their search space from elites in the MAP-Elites archive. 

Several works in MBQD have drawn inspiration from MBRL. Previous work~\citep{Keller2020ModelBasedQS} proposes a model-based method for the Novelty Search via Local Competition (NSLC)~\citep{lehman2011evolving} algorithm that predicts both the agent reward and a skill descriptor of a reinforcement learning-based robotics controller. Another approach~\citep{Lim2021DynamicsAwareQF} trained a model to predict the environment dynamics of a robot. The dynamics model can then be leveraged to learn robotic skills for downstream tasks. Given that the model predicts the dynamics, rather than reward or skill descriptors, new models can be trained to solve different downstream tasks by defining a reward function for the dynamics model. These approaches specialize for skill learning in reinforcement learning domains. Our approach differs as it can be applied to any optimization problem that can be modeled as a quality diversity problem, assuming that a suitable surrogate model exists for the search space.

\begin{algorithm}[!t]
\caption{DSA-ME}\label{alg:DSA-ME}
\LinesNumbered
\SetKwInput{KwInput}{Input}                %
\SetKwInput{KwOutput}{Output}              %
\DontPrintSemicolon
  \KwInput
  {\textbf{\textit{evaluate}}: function that returns objective $f$, measures $\mathbf{m}$, and ancillary data $\mathbf{\alpha}$ \newline
  \textbf{\textit{random\_solution}}: function that generates a random solution $x$\newline
  \textbf{\textit{perturb\_elite}}: function that generates a new solution $x$ given an input archive\newline
  $\textit{\textbf{N}}$: Number of evaluations on \textit{evaluate} function\newline
  $\textit{\textbf{n}}$: Number of inner loop iterations\newline
  $\text\it{\textbf{G}}$: Initial population size}
  \KwOutput{Ground-truth Archive $M$}

Initialize a ground-truth archive $M$, a surrogate model $sm$, and a data buffer $D$ \\
    $i \gets 0$ \\
    \While{$i < N$}{
        \eIf{$i < G$}
        {
            $x \gets$ \textit{random\_solution()} \\
            $f, \mathbf{m}, \mathbf{\alpha} \gets$ \textit{evaluate($x$)} \\
            Add ($x$, $f$, $\mathbf{m}$, $\mathbf{\alpha}$) to $D$ \\
            Add ($x$, $f$, $\mathbf{m}$, $\mathbf{\alpha}$) to $M$ \\
            $ i \gets i + 1$ \\
        }{
            $sm.train(D)$ \\
            Initialize a surrogate archive $M_s$ \\
            \For{$j \gets 0$ \KwTo $n$}{
                \eIf{$j < G$}
                {
                    $x \gets$ \textit{random\_solution()} \\
                }
                {
                    $x \gets$ \textit{perturb\_elite($M_s$)} \\
                }
                $\hat{f}, \hat{\mathbf{m}}, \mathbf{\hat{\alpha}}  \gets sm.predict(x)$ \\
                Add ($x$, $\hat{f}$, $\hat{\mathbf{m}}, \mathbf{\hat{\alpha}}$) to $M_s$ \\
            }
            \For{$x \in M_s$}{
                $f, \mathbf{m}, \mathbf{\alpha} \gets$ \textit{evaluate($x$)} \\
                Add ($x$, $f$, $\mathbf{m}$, $\mathbf{\alpha}$) to $D$ \\
                Add ($x$, $f$, $\mathbf{m}$, $\mathbf{\alpha}$) to $M$ \\
                $ i \gets i + 1$ \\
            }
        }
    }
\end{algorithm}

\section{Approach: Deep Surrogate \\Assisted MAP-Elites}

We introduce the Deep Surrogate Assisted MAP-Elites (DSA-ME) algorithm, a model-based quality diversity algorithm derived from MAP-Elites. Fig~\ref{dsail} gives an overview of the DSA-ME framework and Algorithm~\ref{alg:DSA-ME} provides a pseudocode implementation of DSA-ME.

\subsection{Deep Surrogate Model}\label{smodel}

In our Hearthstone example domain, a deep surrogate model predicts gameplay outcomes for an input Hearthstone deck. The model acts as an efficient surrogate for the game engine, predicting the objective $f$ and measures $\textbf{m}$ needed to run MAP-Elites. By evaluating game content on the surrogate model instead of the game engine, we aim to improve the efficiency of generating high quality and diverse decks. Additionally, we explore a surrogate model that predicts ancillary data $\alpha$ extracted from gameplay, hypothesizing that  this helps form better internal representations and thus generalizes better to new decks.

\subsection{Inner and Outer Loops}

The DSA-ME algorithm consists of an inner and outer loop visualized in Fig~\ref{dsail}. The inner loop exploits the surrogate model to discover a diverse archive of high-quality solutions, \emph{according to the surrogate model}. The outer loop evaluates candidate archives produced by the inner loop on the Hearthstone game engine as a way to (1) generate labeled data to improve the surrogate model and (2) update a running ground-truth archive of high-quality decks with diverse strategic gameplay.

We initialize the outer loop in the same manner as MAP-Elites, by generating $G$ random decks and evaluating them on the Hearthstone game engine. These initial decks as well as their gameplay data form the initial training data $D$ for the surrogate model and the initial solutions for the outer archive $M$. 

After generating the initial $G$ decks and training the initial surrogate model (line 11), we initialize a new archive $M_s$ (line 12) containing decks with gameplay outcomes predicted by the surrogate model $sm$. As in MAP-Elites, the inner loop generates the first $G$ decks by randomly sampling decks (line 15) and later decks by perturbing existing elites (line 16). Within the inner loop, we evaluate new candidate decks directly on the surrogate model (line 19). After evaluation, each candidate deck $x$ and surrogate predictions $\hat{f}$, $\mathbf{\hat{m}}$, and $\hat{\alpha}$ gets added to the surrogate archive $M_s$ (line 20).

After the inner-loop runs $n$ iterations, we return to the outer loop. We extract all elites from the surrogate archive $M_s$ and evaluate them on the Hearthstone to obtain ground-truth predictions $f$, $\mathbf{m}$, and $\alpha$ (lines 22-27). These decks and their ground-truth data get added to both the training data buffer for the surrogate model (line 24) and the outer loop's ground truth archive (line 25). The algorithm results in a ground-truth archive $M$ with decks that have been evaluated on the Hearthstone game engine.

\subsection{On the Benefits of  Online Training} 

In this section, we provide our intuition on why we expect online training of the surrogate model to improve search performance over surrogate models trained from offline data.

The initial surrogate model will struggle to predict gameplay outcomes as the model is trained only on a few randomly chosen initial decks. When we run MAP-Elites to exploit the surrogate model in the inner loop, we will discover areas of the search space where the surrogate model incorrectly predicts the objective or measure functions of a given deck. Moreover, the data extracted from the surrogate model is diverse according to the surrogate model. By training the surrogate model on these adversarial, with respect to the objective, yet diverse solutions, the model will update its predictions on what constitutes a good Hearthstone deck.

As the outer loop progresses, the surrogate model will become more accurate. When the inner loop starts generating more accurate solutions, the surrogate archive will provide better estimates for high quality decks that vary according to the measure functions. Gradually, the role of the inner loop \mbox{MAP-Elites} shifts from generating adversarial decks that exploit the surrogate model to generating high quality decks that populate the ground-truth archive.

 We contrast the online approach with surrogate models trained only with offline data. In the offline case, MAP-Elites will exploit the surrogate model for decks that the surrogate model predicts will perform well, but that perform poorly when evaluated on Hearthstone. Without including these adversarial decks in the training set, the model cannot correct its erroneous predictions.

\section{Experiments} \label{sec:exp}
Our experiments show that combining the QD algorithm MAP-Elites with a \textit{deep surrogate} model trained \textit{online} improves the quality and diversity of automatically generated Hearthstone decks.

\subsection{Experiment Setup}

\subsubsection{Search Space}
We search for decks for the \textit{Rogue} class in the basic and classic card sets. We are interested in \textit{Rogue} because decks in this class require long-term planning due to the class's combo mechanic; some cards played in the early states of the game might be useful in the later states when played with other cards~\cite{Fontaine_CMAME_2020}. Each deck contains 30 cards, selected from 180 candidate cards. Each card can be selected at most twice.

To input a deck into the surrogate model, we encode the deck as an integer vector that represents the frequency of each card chosen. Formally, each input vector $x \in \mathbb{R}^c$ represents a deck where $c$ is the cardinality of the cardset. Each element $x_i$ represents the number of copies of card $i$ included in the deck. The range of $x_i$ is bounded to $\{0, 1, 2\}$ because each card can be added to a deck at most twice in Hearthstone. We call this encoding method \textit{Bag-of-Cards} as it is similar to the Bag-of-Words~\cite{zhang2010understanding} embedding used in natural language processing and information retrieval problems.

\subsubsection{SabberStone Simulator}
To simulate games of Hearthstone, we run our experiments on SabberStone~\cite{SabberStone}, a community developed simulator of Hearthstone games. %

\subsubsection{Objective and Measures} \label{subsec:obj_meas} We adopt the same objective and measures proposed by previous work~\citep{Fontaine_CMAME_2020} for both strategy training and deck search. To evaluate a deck or strategy, we play 200 games of Hearthstone distributed evenly among our opponent suite. To provide a smooth approximation of winning percentage, we specify an objective of the average health difference between the two players at the end of each game. 
For measures, we calculate the average number of cards in hand at the start of each turn and the average number of turns across the 200 games. The average number of turns measure defines the spectrum between \textit{aggro}, which ends the game quickly, and \textit{control}, which tries to prolong the game, playstyles.

\subsubsection{Player Strategy}

We use CMA-ME~\cite{Fontaine_CMAME_2020} to train a single high-performing neural network to play a human-designed \textit{Miracle Rogue} deck~\citep{classic_miracle_rogue}  with the objective and measures defined in section~\ref{subsec:obj_meas}. We provide more detail on the player strategy in Appendix~\ref{subsec:player_strategy}. After training, we evaluate the generated decks using the same \textit{fixed} player strategy for all tested algorithms.

\subsubsection{Opponents}
We selected six high-performing decks discovered by previous work~\citep{fontaine:gecco19} in the \textit{Hunter}, \textit{Paladin}, and \textit{Warlock} classes as our opponents. For each class, we include both an \textit{aggro} and \textit{control} opponent. We choose these opponents to provide our deck search with high quality and strategically diverse opponents.

\subsubsection{Surrogate Models and Baselines} We use a multilayer perceptron (MLP) as a deep surrogate model. To show the benefit of deep surrogate models, we include a linear surrogate model baseline. We refer to these models as Linear Surrogate Assisted MAP-Elites (LSA-ME) and Deep Surrogate Assisted MAP-Elites (DSA-ME), respectively. To demonstrate the importance of training the surrogate model online, we add a baseline that exploits a fixed MLP trained in a supervised fashion. We collect the training data for the offline MLP by sampling 10000 randomly generated decks and evaluating them on SabberStone. The fixed model is not updated online after training. We denote this baseline as \fixedfcnneme. Finally, we include MAP-Elites without a surrogate model as a model-free baseline. We run 5 trials for each algorithm. We describe the hyperparameters in Appendix~\ref{subsec:Hyperparameters}.

\subsubsection{Perturbation Operator.} All implemented QD algorithms require a perturbation operator. To perturb elites, we replace $k$ cards from a parent elite with $k$ cards chosen uniformly from the cardset. We vary $k$ geometrically, following previous work~\citep{fontaine:gecco19}.

\subsubsection{Ancillary Data} \label{subsec:am}

Drawing upon insights from self-supervised learning~\citep{ren2018cross}, we explore whether adding ancillary data to the prediction of the surrogate models helps form better internal representations and thus generalize better to new decks. As ancillary data, we include the win percentage, the total damage done, the number of cards drawn, the total mana spent, the total mana wasted, the sum of mana cost for all cards in the deck, the variance of mana costs in the deck, the number of minion cards in the deck, and the number of spell cards in the deck. We chose data that relates to either deck composition or how the player strategy utilized the given deck. We denote surrogate model algorithms trained with ancillary data as DSA-ME (ad) and Offline DSA-ME (ad).

\subsection{Experiment Design}

\subsubsection{Independent Variables} 
We follow a between-groups design with two independent variables: (1) The type of surrogate model: DSA-ME, LSA-ME, Offline DSA-ME, and MAP-Elites without any surrogate model. (2) The presence of ancillary data: none or ancillary data (ad) if the model predicts the additional metrics described in section~\ref{subsec:am}.

\subsubsection{Dependent Variables}
The dependent variables measure the quality and diversity of the decks found by the algorithms. Specifically, the dependent variables are the maximum average health difference, the maximum win percentage, the percentage of cells filled in the ground-truth archive, and the QD-score~\citep{Justin2015confronting}. The maximum average health difference and maximum win percentage measure the capability of the algorithm to search for globally optimal solutions. The percentage of cells filled in the ground-truth archive measures the diversity of the solutions. The QD-score measures the quality and diversity of the solutions altogether and is defined as the sum of the objective values of all solutions in the ground-truth archive.

\subsubsection{Hypotheses}

\ \newline
\noindent\textbf{H1:} \textit{The \fcnneme, \lineareme, and \fixedfcnneme\ algorithms will have better performance than \mapelites}. We hypothesize that the surrogate models will result in MAP-Elites finding more diverse and high-performing solutions than without the models.

\noindent\textbf{H2:} \textit{ \fcnneme\ will have better performance than  \fixedfcnneme\ and \lineareme}. We choose as baseline the linear model to demonstrate the benefit of a deep model. We also compare against \fixedfcnneme\ to demonstrate the benefit of training the model online with data generated from MAP-Elites.

 \noindent\textbf{H3:} \textit{Adding the ancillary data to DSA-ME and Offline DSA-ME will improve performance}. We expect the ancillary data to give an additional training signal to the surrogate models so that the models can learn a better internal representation and more accurately predict the objective and measures. We only include ancillary data in DSA-ME and Offline DSA-ME, since linear models do not have internal representations.

\begin{table*}[h]
\centering
\begin{tabular}{l|c|c|c|c} 
    \hline
    Algorithm & Max Avg Health Diff & Max Win Percentage & Cells Filled Percentage & QD-Score \\ [0.5ex] 
    \hline
    \mapelites & 15.72 $\pm$ 0.62 & 90.6 $\pm$ 0.91 \% & 14.36 $\pm$ 0.32 \% & 136.89 $\pm$ 1.95 \\ 
    \fixedfcnneme & 18.69 $\pm$ 0.17 & 97.1 $\pm$ 0.19 \% & 21.64 $\pm$ 0.12 \% & 194.94 $\pm$ 1.65 \\
    \lineareme & 20.07 $\pm$ 0.18 & 96.2 $\pm$ 0.12 \% & 27.89 $\pm$ 0.64 \% & 276.23 $\pm$ 6.10 \\
    \textbf{\fcnneme} & \textbf{22.30 $\pm$ 0.16} & \textbf{98.4 $\pm$ 0.10 \%} & \textbf{31.86 $\pm$ 0.71 \%} & \textbf{338.52 $\pm$ 5.86} \\
    \fixedfcnnememt & 20.23 $\pm$ 0.20 & 97.7 $\pm$ 0.30 \% & 15.65 $\pm$ 0.13 \% & 168.67 $\pm$ 0.99 \\
    \fcnnememt & 21.15 $\pm$ 0.35 & 97.4 $\pm$ 0.43 \% & 30.78 $\pm$ 1.35 \% & 312.66 $\pm$ 11.72 \\
    \hline
\end{tabular}
  \caption{Hearthstone numerical results}
  \label{numerical}

\end{table*}

\begin{figure*}[!ht]
\vspace{-1em}
  \centering
  \subfigure{\includegraphics[scale=0.25]{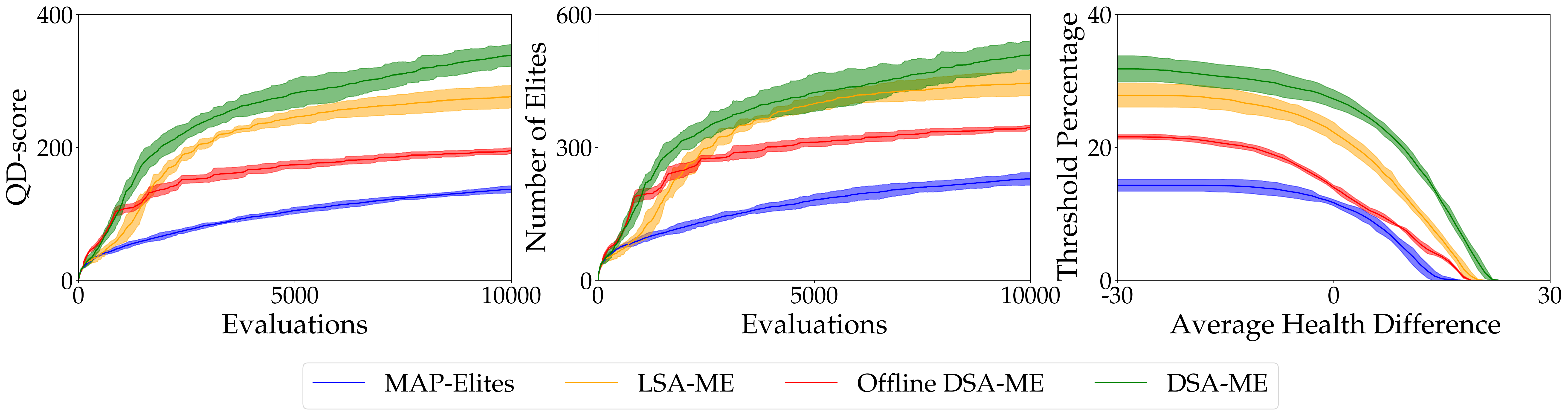}}\quad
  \caption{Quantitative analysis of the elites in the archive with the surrogate model predicting objective and measures (without ancillary data). In the rightmost plots, the y-axis shows the percentage of filled cells in the archive that had average health difference larger than the corresponding value in the x-axis. Larger area under each curve indicates better performance}
  \label{analysis}
\end{figure*}

\begin{figure*}[t!]
\centering
\includegraphics[width=1.0\linewidth]{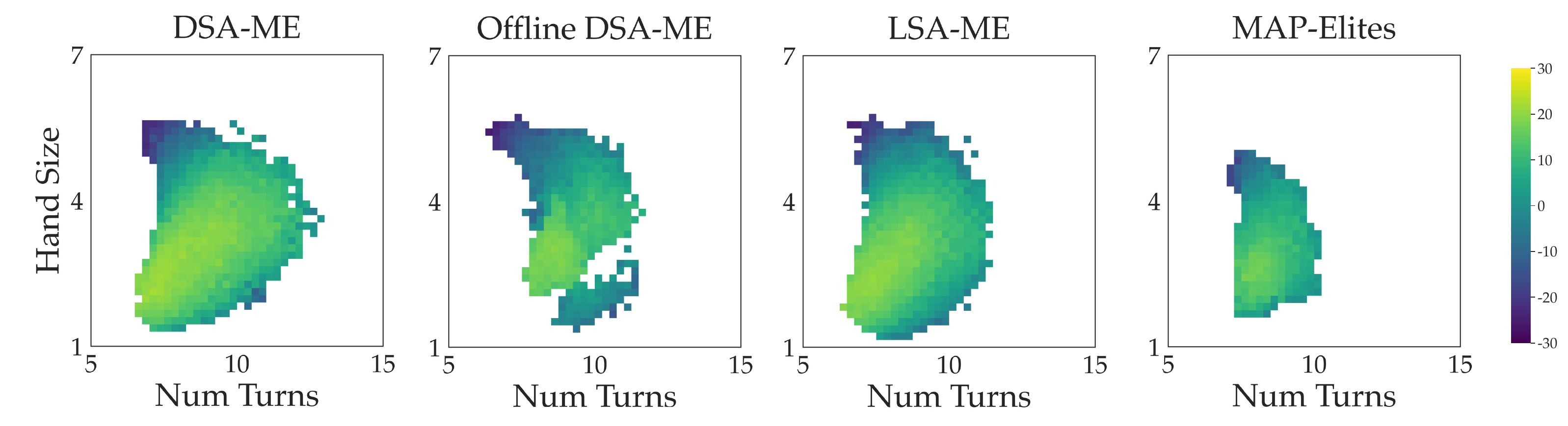}
\vspace{-1em}
\caption{Example archives from single runs of vanilla MAP-Elites and variants of DSA-ME. The color indicates the objective value (average
health difference) of each solution. Qualitatively DSA-ME finds more diverse and high performing solutions than 
MAP-Elites.}
\label{heatmap}
\end{figure*}

\section{Results} \label{sec:results}

Table~\ref{numerical} shows the average maximum objective value, average maximum win percentage, average percentage of cells filled, and average QD-score over 5 trials. %

To test hypotheses 1 and 2, we performed a one-way ANOVA test for each of the metrics. There were statistically significant differences in all four metrics: QD-score ($\textit{F}(3, 16)=403.49, \textit{p}<0.001$), maximum win percentage ($\textit{F}(3, 16)=52.81, \textit{p}<0.001$), maximum average health difference ($\textit{F}(3, 16)=64.65, \textit{p}<0.001$), and percentage of cells filled ($\textit{F}(3, 16)=228.30, \textit{p}<0.001$).  Post-hoc pairwise comparisons with Bonferroni corrections showed that all three variations of DSA-ME performed significantly better than \mapelites\ in all four metrics. This supports \textbf{H1}.

In addition, \fcnneme\ outperforms \lineareme\ and \fixedfcnneme\ in QD-score, percentage of cells filled, and maximum average health difference. There is no statistically significant difference in the maximum win percentage among the three algorithms. We attribute this to the fact that the maximum win percentage is high for all algorithms. Overall, we observe that DSA-ME finds more diverse and high-performing solutions, supporting \textbf{H2}. 

To test the third hypothesis, we first performed a two-way ANOVA test for each metric, with the type of the surrogate model (DSA-ME, Offline DSA-ME) and the presence of ancillary data as independent variables. There was no statistically significant interaction between the surrogate model type and the presence of ancillary data for QD-score. Main effect analysis showed that surrogate models with ancillary data had significantly smaller QD-score than the models without the ancillary data ($p=0.001$).

We found a statistically significant interaction between the two independent variables for maximum win percentage ($\textit{F}(1, 16)=8.00, \textit{p}=0.012$), maximum average health difference ($\textit{F}(1, 16)=33.277, \textit{p}<0.001$), and percentage of cells filled ($\textit{F}(1, 16)=10.148, \textit{p}=0.006$). 
Simple main effect analysis showed that ancillary data helped only in one case: maximum average health difference was significantly higher for \fixedfcnnememt\ compared to \fixedfcnneme\ ($p<0.001$). For \fcnnememt, maximum win percentage and maximum average health difference are both significantly worse  than \fcnneme\ ($p=0.024,p=0.003$) and we found no significance in percentage of cells filled. For \fixedfcnnememt, percentage of cells filled is significantly worse than \fixedfcnneme\ ($p<0.001$) and we found no significance for maximum win percentage. These results show that adding ancillary data did not improve performance and they do not support \textbf{H3}. We discuss this finding in section~\ref{sec:discussion}.

Figure~\ref{analysis} shows the QD-score and the percentage of cells filled. We note that the QD-score metric combines both the number of cells filled in the archive (diversity) and the objective value of each occupant (quality). To disambiguate the two, we also show in Fig.~\ref{analysis} the  Complementary Cumulative Distribution Function (CCDF) plot of the solutions in the archive. The CCDF plot shows the percentage of cells (y-axis) that have objective value greater than the threshold specified in the x-axis. Fig.~\ref{heatmap} shows example heatmaps of the ground-truth archives of the four algorithms (without the ancillary data).

The heatmaps and plots support the results from the statistical analysis, showing that DSA-ME finds more diverse and higher quality solutions than the baselines. They also show that \fixedfcnneme, although using a deep surrogate model, performs worse than \lineareme\ in both QD-score and number of elites. In addition, as shown in the CCDF plots, \fcnneme\ consistently finds more solutions in the archive over all average health differences.

\begin{figure*}[!ht]
  \centering
  \subfigure{\includegraphics[scale=0.25]{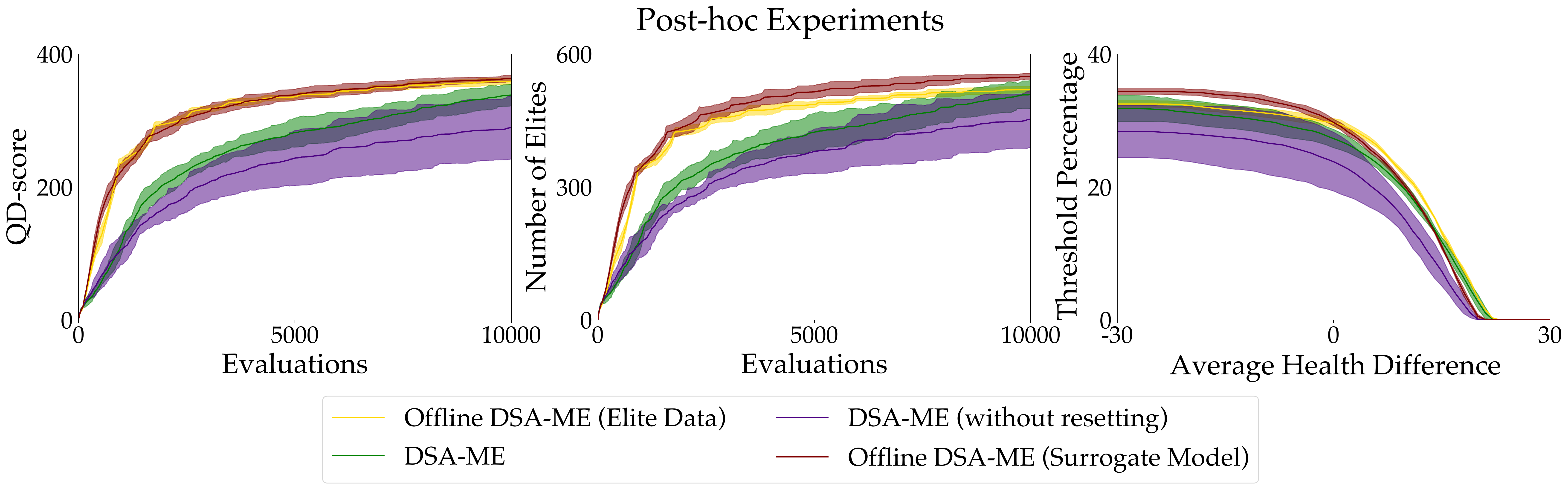}}
  \vspace{-1em}
  \caption{ Post-hoc experiments to evaluate the quality of the surrogate model and the effect of resetting MAP-Elites in the inner loop of DSA-ME.}
  \label{fig:post-hoc-analysis}
\end{figure*}

\section{On the Performance of DSA-ME} 
We perform a post-hoc experimental analysis to shed more light on what makes DSA-ME achieve high performance. 

\subsection{On the Quality of the Surrogate Model} \label{subsec:quality_model} 
First, we analyze the quality of the surrogate model to determine if high quality predictions of the objective and measure functions result in the improved performance. Consider instead an alternative hypothesis that the surrogate model \emph{drifts} in prediction error and the fact the surrogate model changes throughout DSA-ME is responsible for good performance rather than prediction accuracy.

First, we confirm that the final trained surrogate model from DSA-ME achieves high accuracy by evaluating it on a different test data created from a separate run of DSA-ME. The average mean squared error over 5 runs is 0.44 for the predicted number of turns whose range is [5, 15], 0.07 for the hand size with range [1, 7], and 14.70 for the average health difference with range [-30, 30]. We observe that the model predicts the measure function outputs with high accuracy, while it has higher error for the objective.

\begin{figure}[!t]
  \centering
\includegraphics[scale=0.25]{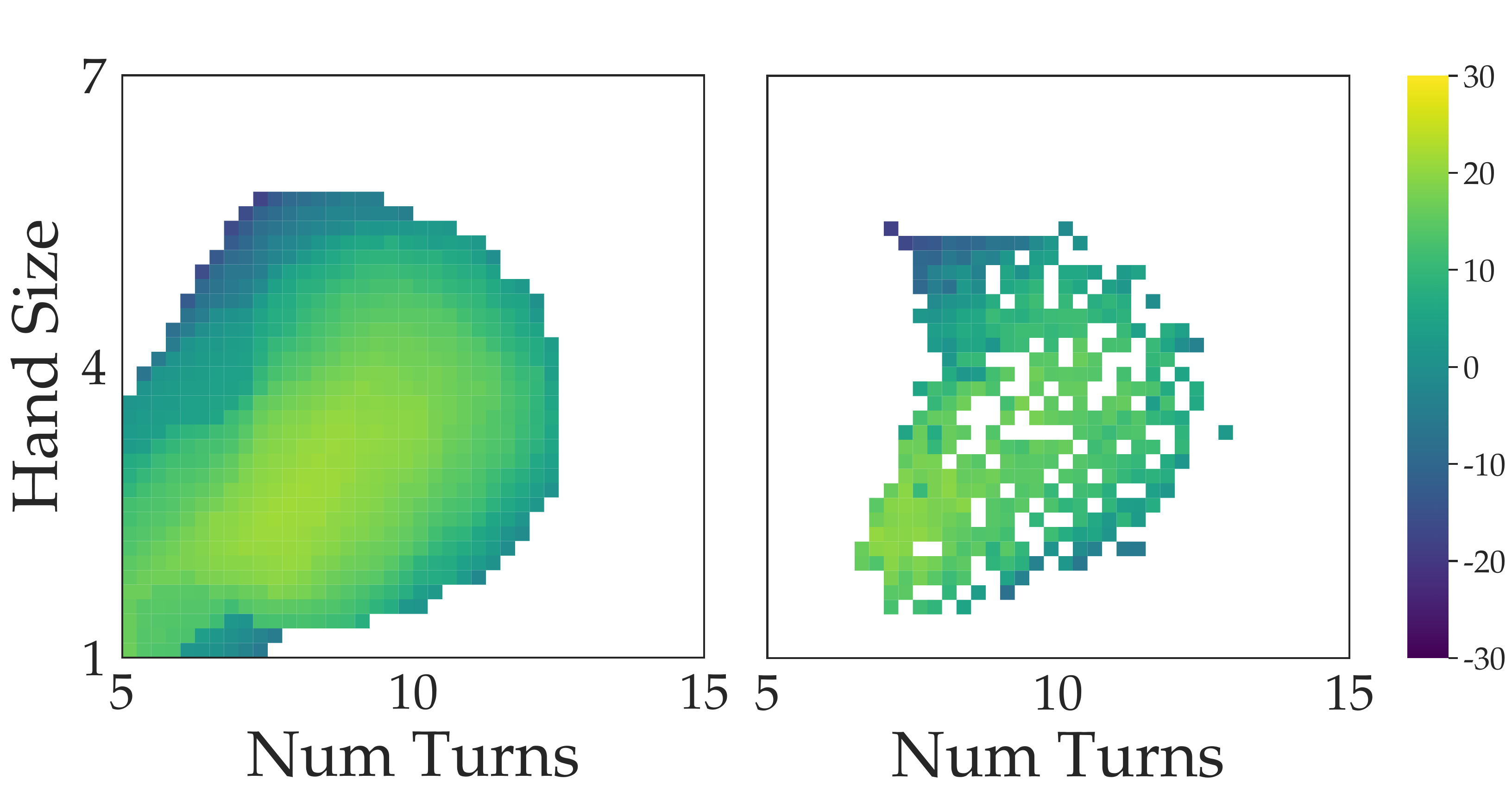}
  \caption{(Left) An archive produced by a single run of MAP-Elites on the surrogate model. (Right) An archive produced by evaluating the elites on the left with the game engine.}
  \label{fig:surrogate_real}
\end{figure}

While the above measure function predictions are accurate, they are not accurate enough to ensure the candidate solutions from the surrogate archive will remain in the same MAP-Elites bin when evaluated on the game engine. To evaluate this, we run MAP-Elites on the final DSA-ME surrogate model for 5 trials. We then take elites from the surrogate archive, evaluate each deck on the game engine, and populate a new archive with the ground truth objective and measures (see Fig.~\ref{fig:surrogate_real}). We observe that 44\% of the elites from the surrogate archive are retained in the new archive. While only 3\% of decks are placed in the same cell as the surrogate archive, the average Manhattan distance between the cell in the surrogate and ground-truth archives is 2.45 cells. This result shows that although the surrogate model does not frequently predict the exact cell, candidate solutions remain in the same region of the archive.

To show that the quality predictions of the surrogate model are the reason for the high performance of DSA-ME, we add as baseline Offline DSA-ME (Surrogate Model), which uses a ``frozen'' surrogate model that is fully trained from a previous run of DSA-ME. Fig.~\ref{fig:post-hoc-analysis} shows that Offline \mbox{DSA-ME} (Surrogate Model) outperforms \mbox{DSA-ME}. This indicates that using a fully-trained model for evaluation results in better performance than training the model online. We note that \textit{this holds only when the surrogate model has been trained with data generated from a complete DSA-ME run}; if it is trained with randomly generated data, Offline DSA-ME performs significantly worse, as shown in section~\ref{sec:results}. We include an additional baseline,  Offline DSA-ME (Elite Data) described in Appendix~\ref{subsec:quality_data}.

Overall, results show that \textit{running MAP-Elites in the inner-loop of Algorithm 1 results in a diverse, high-quality dataset of decks for the surrogate model. Training on the dataset enables the surrogate model to make accurate predictions}.

\subsection{On the Importance of Resetting MAP-Elites} \label{subsec:reset}

We visualize the archive generated by running MAP-Elites using the surrogate model predictions (Fig.~\ref{fig:surrogate_real}-left). We then evaluate each elite on the Sabberstone game engine and use the Sabberstone evaluations to place the elites on a new archive (Fig.~\ref{fig:surrogate_real}-right). We observe that the new archive covers the space reasonably well but has holes in different areas.

Thus, we hypothesize that resetting MAP-Elites in the inner-loop (line 12) with random initialization produces archives with different missing cells and that the union of these archives covers the space well. We test this hypothesis by running DSA-ME and keeping the same archive in the inner-loop, instead of initializing a new archive (lines 12-18 in Algorithm 1). We call this baseline DSA-ME (without resetting). Fig.~\ref{fig:post-hoc-analysis} shows that this baseline has lower QD-score and coverage. %
We find this result consistent with previous work~\cite{lehman2015enhancing,veenstra2020death} on the benefits of mortality on evolvability in divergent search algorithms.

\section{Discussion} \label{sec:discussion}

Our work presents a method to efficiently generate diverse and high-quality game content for collectable card games. We introduce the Deep Surrogate Assisted MAP-Elites (DSA-ME) algorithm, a novel model-based quality diversity algorithm. 

In our experiments, we observe that DSA-ME outperforms surrogate models trained on offline, randomly generated data and a model-free MAP-Elites approach, setting a new state-of-the-art for QD approaches in automated Hearthstone deckbuilding. By dynamically forming a surrogate model of Hearthstone gameplay, DSA-ME explores more complex decks, whose performance is hypothesized by surrogate model predictions. This reduces the number of simulated games required to find high-performing decks with differing strategic gameplay. We anticipate that our results will benefit game designers looking to balance collectable card games through automated deckbuilding~\citep{demesentier:cog19}.

While our DSA-ME approach improves upon existing methods for automated deckbuilding, we found that adding ancillary data to the surrogate model predictions did not improve performance. This result conflicts with our intuitions from recent advancements in self-supervised learning~\citep{ren2018cross}; we hypothesized that ancillary data would help the deep surrogate model learn a better internal representation of the Hearthstone cardset.

A potential reason for this result is the simplicity of the Hearthstone deck representation. We encode decks as a Bag-of-Cards input feature vector, which marks if a card is included in the deck.  This encoding does not capture a card's mana cost, attack, health, or card text that would be available to a human player when building a deck. We conjecture that encoding these card attributes could help the surrogate model form a better assessment of individual cards and generalize better to new cards.

An alternative reason could be the low fidelity of our 
surrogate model. A low fidelity model would predict only game outcomes, while a high fidelity model would replicate the entire game engine. 
Low-fidelity models allow for game content to be evaluated efficiently as a single forward pass in a feed-forward neural network.  However, the low fidelity model may not be able to represent the level of context necessary to take advantage of ancillary data.

An alternative approach would be to use a high-fidelity model that predicts the game dynamics directly, similarly to model-based reinforcement learning~\cite{Lim2021DynamicsAwareQF}. A high-fidelity neural simulation of Hearthstone games could help improve the prediction accuracy of the surrogate model. However, such a model could be less efficient than the Hearthstone simulator itself. Future work will explore this trade-off across the fidelity spectrum of model-based quality diversity for procedural content generation.

\begin{acks}
The authors thank Adam Gaier for his  feedback in early versions of this work. 
\end{acks}

\bibliographystyle{IEEEtranN}
\bibliography{references}
\clearpage
\appendix

\section{Implementation Details}
\subsection{Hyperparameters}
\label{subsec:Hyperparameters}

The MLP of the surrogate model has three hidden layers with sizes $[128, 32, 16]$ with Exponential Linear Unit (ELU) activation functions. All surrogate models are trained using the mean squared error loss function and Adam optimizer~\cite{kingma2014adam} with learning rate $\eta = 0.01$ and batch size 64. In each outer loop, all models are trained for 20 epochs. 

We set the number of inner loop iterations $n = 10^6$ with a batch size of 10. We set the number of evaluations on the SabberStone game simulator $N = 10,000$, each with 200 simulated Hearthstone games. The initial population size $G$ for both inner and outer loop is 100. We run all algorithms for 5 trials on a high-performing cluster, utilizing 200 Xeon CPUs. Each trial lasted 10 to 12 hours.

\subsection{Player Strategy} \label{subsec:player_strategy}

SabberStone includes a library containing simple AI agents that play the game via a tree search. At each turn,  the tree search explores possible moves and discovers the best action sequence that maximizes the heuristic game state score at the end of the turn. Included heuristics emulate \textit{aggro} strategies, which end the game quickly,
and \textit{control} strategies, which try to prolong the game, for arbitrary decks. The library allows for custom scoring functions that result in more advanced strategies~\citep{Fontaine_CMAME_2020}.

In our preliminary experiments, we found that the heuristic strategies included in SabberStone struggle to play decks of the \textit{Rogue} class. To create a stronger agent, we train a custom neural network heuristic function with \mbox{CMA-ME} following previous work~\citep{Fontaine_CMAME_2020}. The neural network maps 15 game related features defined by SabberStone~\citep{SabberStone} to a single value representing the preference for an end-of-turn state. The neural network is a multilayer perceptron with layer sizes $[15, 5, 4, 1]$ and 109 parameters.

We train the neural network to play a human designed \textit{Miracle Rogue} deck~\citep{classic_miracle_rogue} as the player's deck with the objective and measures defined in section~\ref{subsec:obj_meas}. The \textit{Classic Miracle Rogue} deck requires the player to have long-term planning to play well, thus we expect a high-performing strategy using this deck to generalize to other advanced \textit{Rogue} decks as well.

While CMA-ES~\citep{hansen2016cmaes} is usually preferred while optimizing a single objective black-box function, previous work~\citep{Fontaine_CMAME_2020} showed that \mbox{CMA-ME} is able to find a globally better neural network strategy than CMA-ES because CMA-ME finds more diverse solutions and is less likely to converge to a local minimum. Therefore, the neural network is trained using CMA-ME in 50000 evaluations. The highest performing solution was chosen as the final strategy.

\section{On the Quality of the Labeled Elite Data} \label{subsec:quality_data} 

We note the similarity of training the surrogate archive to the online training of the discriminator network in generative adversarial networks~\cite{goodfellow2014generative}. We test whether we can only train the surrogate model incrementally in a closed-loop manner, or we can train it from scratch using the entire dataset generated from a DSA-ME run. We compare the final surrogate model of a DSA-ME run, Offline DSA-ME (Surrogate Model), with a surrogate model trained from scratch on the data of the DSA-ME run, Offline DSA-ME (Elite Data). In both conditions we train the surrogate models and then run DSA-ME with the frozen models. Fig.~\ref{fig:post-hoc-analysis} shows that there is no significant difference in performance between the two models, indicating that we can use the training data from a DSA-ME run to train an accurate surrogate model from scratch.

\end{document}